\documentclass[letterpaper, 10 pt, conference]{ieeeconf}

\usepackage{amsmath}
\usepackage{amssymb}
\usepackage{graphicx}
\usepackage{siunitx}
\usepackage{pbox}

\usepackage[shortcuts]{extdash}
\usepackage{ulem}

\usepackage{bm}

\usepackage[usenames,dvipsnames]{xcolor}

\newcommand{\replace}[2]{%
\ifthenelse{ \equal{#1}{} }{}{\textcolor{olive}{\sout{#1}}}%
\ifthenelse{ \equal{#2}{} }{}{ \textcolor{olive}{#2}}%
} 

\newcommand{\revised}[1]{\textcolor{black}{#1}}

\title{\LARGE \bf Generation of GelSight Tactile Images\\ for Sim2Real Learning}
\author{Daniel Fernandes Gomes$^{1}$, Paolo Paoletti$^{2}$ and Shan Luo$^{1}$
\thanks{$^{1}$smARTLab, Department of Computer Science, University of Liverpool, Liverpool L69 3BX, United Kingdom.}
\thanks{$^{2}$School of Engineering, University of Liverpool, Liverpool L69 3GH, United Kingdom.}
\thanks{Emails: \tt\{danfergo, paoletti, shan.luo\}@liverpool.ac.uk}}

\begin{document}

\maketitle
\thispagestyle{empty}
\pagestyle{empty}

\begin{abstract}
Most current works in \textit{Sim2Real} learning for robotic manipulation tasks leverage camera vision that may be significantly occluded by robot hands during the manipulation. Tactile sensing offers complementary information to vision and can compensate for the information loss caused by the occlusion. However, the use of tactile sensing is restricted in the \textit{Sim2Real} research due to no simulated tactile sensors being available. To mitigate the gap, we introduce a novel approach for simulating a GelSight tactile sensor in the commonly used Gazebo simulator. Similar to the real GelSight sensor, the simulated sensor can produce high-resolution images by an optical sensor from the interaction between the touched object and an opaque soft membrane. It can indirectly sense forces, geometry, texture and other properties of the object and enables \textit{Sim2Real} learning with tactile sensing. Preliminary experimental results have shown that the simulated sensor could generate realistic outputs similar to the ones captured by a real GelSight sensor. All the materials used in this paper are available at https://danfergo.github.io/gelsight-simulation.

\end{abstract}


\section{Introduction}
The manipulation of objects is prevalent in various applications, e.g., grasping tools, untangling cables and folding pieces of garment. In these tasks, it is essential to track the states of the object being manipulated (shape, pose and the centre of mass etc.). Most methods for robotic manipulation rely on vision-based sensing that allows for a rapid assessment of the scene~\cite{billard2019trends}. However, it can be affected greatly by factors like occlusions and lighting conditions, making the obtained measurements less reliable. In contrast, tactile sensing is not affected by such factors. More importantly, tactile sensing can provide rich contact information between the object and the hand. Nonetheless, tactile sensors are not as widely available as cameras. 

\begin{figure}[t]
\centering
\includegraphics[width=0.49\textwidth]{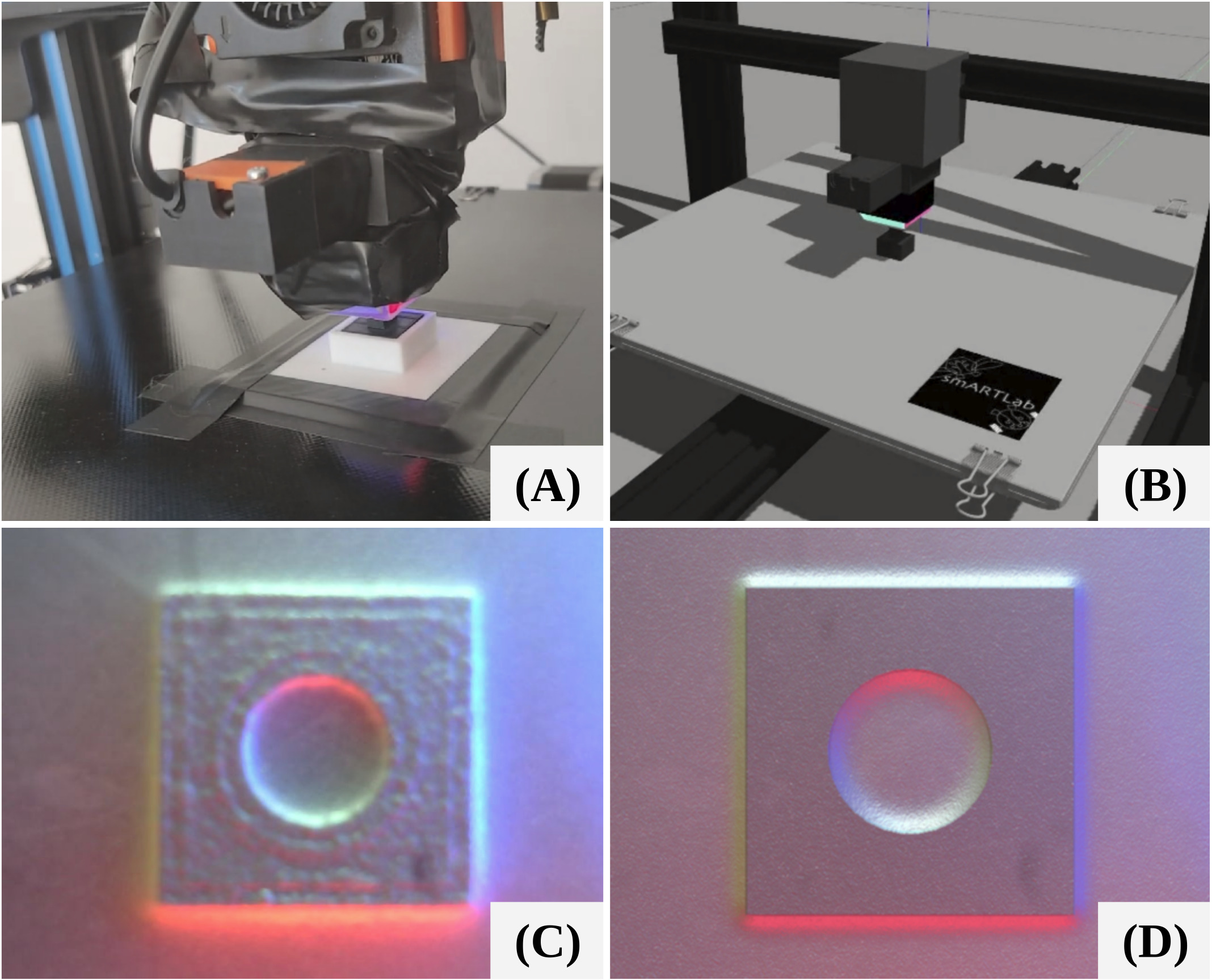}
\caption{\textbf{(A)} and \textbf{(B)}: The real and the simulated experimental setups respectively, in each a GelSight tactile sensor is mounted onto a 3D Printer and contacts a 3D printed object (here is a cube with a hollow cylinder in the centre). \textbf{(C)} and \textbf{(D)}: The corresponding tactile images captured from the real GelSight sensor in (A) and generated from our proposed simulated GelSight sensor in (B) respectively.}
\label{fig:cover}
\end{figure}

To save time and resources, robotic agents can be initially trained within a simulator, so that only a limited number of real experiments are required to fine tune the model before its deployment in the real scenario (\textit{Sim2Real}). A \revised{seminal} example is \cite{DomainRandomization}, where an object detector is pre-trained in randomized simulation environments, and then tested in a real scene. Most of the \textit{Sim2Real} works train models on simulated images that are then transferred to real images. As initial behaviours of the learning agents may be highly unpredictable and can damage delicate tactile sensors, it is also desirable to construct simulated robotic setups equipped with touch sensors, to sense the contact dynamics during manipulation in simulation.  

There have been some works to simulate certain properties of touch sensors, such as friction forces for resistive sensors~\cite{ModelSoftContacts}, surface deformations for optical marker-based tactile sensors~\cite{FEAGreenDots}, and tactile contacts (both detection of contacts and contact locations) for capacitive tactile cells in grasping~\cite{RLRegraspTactileForceFeedback}. However, due to the use of dielectrics~\cite{FEAGreenDots,RLRegraspTactileForceFeedback} or tracking sparse points~\cite{FEAGreenDots,ding2020sim}, these tactile sensors suffer from low resolution, for instance, a commercial Weiss tactile sensor of 14x6 tactile cells simulated in~\cite{ModelSoftContacts}. 

In contrast, GelSight optical tactile sensors~\cite{ GelSightSmallParts, gomes2020blocks} are able to attain high resolution tactile images, thanks to exploiting the full raw images of the elastomer deformation captured by the embedded camera. It has been widely applied to various perception and manipulation tasks, e.g., geometry and slip measurement~\cite{GelSight2017GeometrySlip}, localisation~\cite{GelSlimMappingLocalization,GelSightSmallParts}, texture recognition~\cite{CrossModalTouchVision,ShanVitac}, and tactile servoing~\cite{TactilePredictiveControl}. However, little research has been done on simulating GelSight sensors, which has prevented the exploitation of \textit{Sim2Real} learning for high-resolution tactile sensing.

The main challenge of simulating a GelSight-like sensor is to generate a similar internal view to one captured by the real GelSight camera that depends on the internal illumination and the membrane deformations of the sensor. To overcome the challenge, in this paper we propose a novel approach for simulating a GelSight sensor in the commonly used robotics simulator \textit{Gazebo}\footnote{http://gazebosim.org/}. We leverage a simulated depth camera to capture the surface depth map of the in-contact object. Then, we approximate the heightmap of the deformed membrane by applying Bivariate (2-D) Gaussian filtering. Furthermore, we use the Phong shading model~\cite{phongIlluminationModel} for rendering \revised{the sensor internal illumination.}

To evaluate our proposed method, we collect a dataset of real tactile images using a GelSight sensor~\cite{GelSightSmallParts} and corresponding virtual tactile images, using a set of small 3D printed objects. We use a 3D printer as a Cartesian actuator to perform accurate tapping motions and an equivalent setup is reproduced in the \textit{Gazebo} simulator. Both qualitative and quantitative analyses are performed to compare the difference between real and virtual tactile images, which is as low as 8.39\% on average in the Mean Absolute Error (MAE) that corresponds to a similarity of 0.859 in the Structural Similarity Index Measure (SSIM). To illustrate advantages of using our simulation model, \textit{Sim2Real} learning was conducted to classify a set of 21 objects and the results show that the model trained with only generated tactile images, augmented with random texture perturbations, achieves a high accuracy of 76.19\% when applied to real tactile images.

Our proposed simulation model of the GelSight sensor was firstly presented in the workshop paper~\cite{gomesgelsight} and this paper extends our previous work by including a detailed introduction to the simulation model, rectified Gaussian filtering, and more thorough experiments. Though the \textit{Gazebo} simulator is used in this work, thanks to the simple model and rendering methods we propose, the simulated model can be easily modified and transferred to other simulators such as the Unity\footnote{https://unity.com/} and PyBullet\footnote{https://pybullet.org/wordpress/}.

\section{Related Work}
\label{sec:relatedwork}

\subsection{Simulation of optical tactile sensors}

Tactile sensors of a wide range of working principles have been developed in the last decades~\cite{luo2017robotic,dahiya2009tactile}: resistive,  capacitive, ultrasonic, magnetic, piezo-electric and optical tactile sensors. Compared  to other types, optical tactile sensors, that use cameras to capture the deformations of membranes, have the potential to produce higher resolution tactile images. This advantage is fully exploited by GelSight-like sensors that use the captured raw image, for instance, to reconstruct the contacted surface geometry. In opposition, marker-based optical tactile sensors only track the displacement of sparse markers/pins printed in the soft deformable membrane~\cite{TacTipFamily, ColorMixingTactileSensor, gelForce, GreenDots}. Consequently, the generation of synthetic tactile images depends on the corresponding tactile sensor working principle, especially in simulating the soft membrane physical properties.

It is challenging to simulate the deformation of the elastomer in the contact with another surface. A few approaches take advantage of machine learning algorithms to directly approximate the desired quantities to be measured such as contact forces and incipient slip~\cite{IncipientSlip}. In~\cite{FEAGreenDots}, Finite Element Analysis was used instead to model the deformations of the elastomer. In addition to the distribution of the deformations, the holding torque around the contact surface and the stick-slip phenomenon are modeled using the LuGre dynamic friction model in~\cite{ModelSoftContacts}. For simulating marker-based sensors, the markers displacement can be obtained and tracked~\cite{LearnInSimGreenDots}. The pseudo tactile images can also be generated from real-world data of another modality, for example, visual camera images~\cite{CrossModalTouchVision}. In our work, we use Bivariate Gaussian filtering to generate the protruding surface description, so as to mimic the surface deformation of the elastomer. It is easy to compute and can be used in real-time simulations, which enables fast collection of tactile data in simulation and is ideal for \textit{Sim2Real} applications.

\subsection{Sim2Real Transfer Learning}

Data-driven (or learning-based) approaches, introduce two main advantages over handcrafted ones: require less domain-specific knowledge to develop, and have the potential of continuous online improvement. However, one of their major drawbacks is the requirement of extensive amount of data that is often costly to obtain. To address this issue, Transfer Learning has been proposed: a model pre-trained on a more general domain is fine-tuned to the target domain. It has been used in~\cite{TransferCalibrationLayerGreenDots} to learn a sensor-specific calibration layer, instead of learning the entire model for every sensor. In the context of robotics, it has evolved into \textit{Sim2Real} transfer learning, i.e., training an agent in a simulated environment, followed by its fine-tuning in the real environment, as simulated data is cheaper to obtain than real data. For example, in \cite{RLRegraspTactileForceFeedback}, a Barrett Hand mounted on a robot arm is trained via Sim2Real transfer learning to grasp using proprioceptive and tactile feedback (only contact locations are used).


\section{The GelSight Working Principle}
\label{sec:working_principle}

\begin{figure}
\centering
\includegraphics[width=0.49\textwidth]{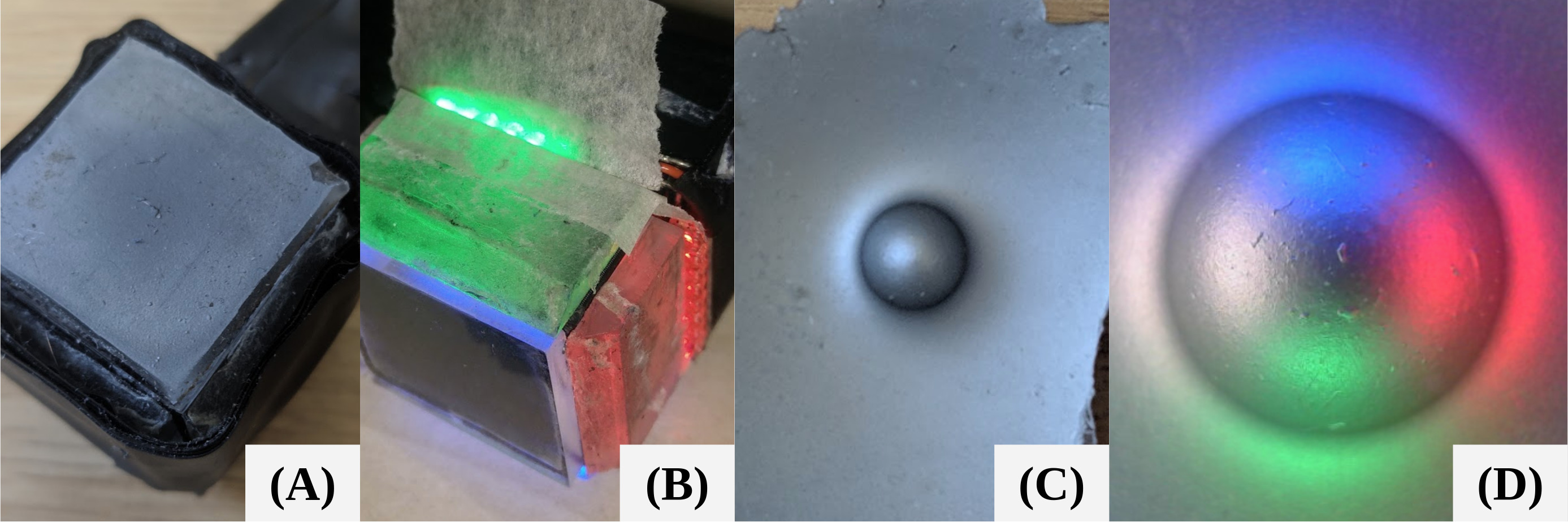}
\caption{Multiple views of the GelSight sensor~\cite{GelSightSmallParts}: \textbf{(A)} the sensor exterior;  \textbf{(B)} the LEDs and the light guiding plates embedded in the sensor; \textbf{(C)} the tactile membrane after being pressed by a ball; \textbf{(D)} the corresponding tactile image captured by the sensor.}
\label{fig:gelsight_working_principle}
\end{figure}

\begin{figure}
\centering
\includegraphics[width=0.35\textwidth]{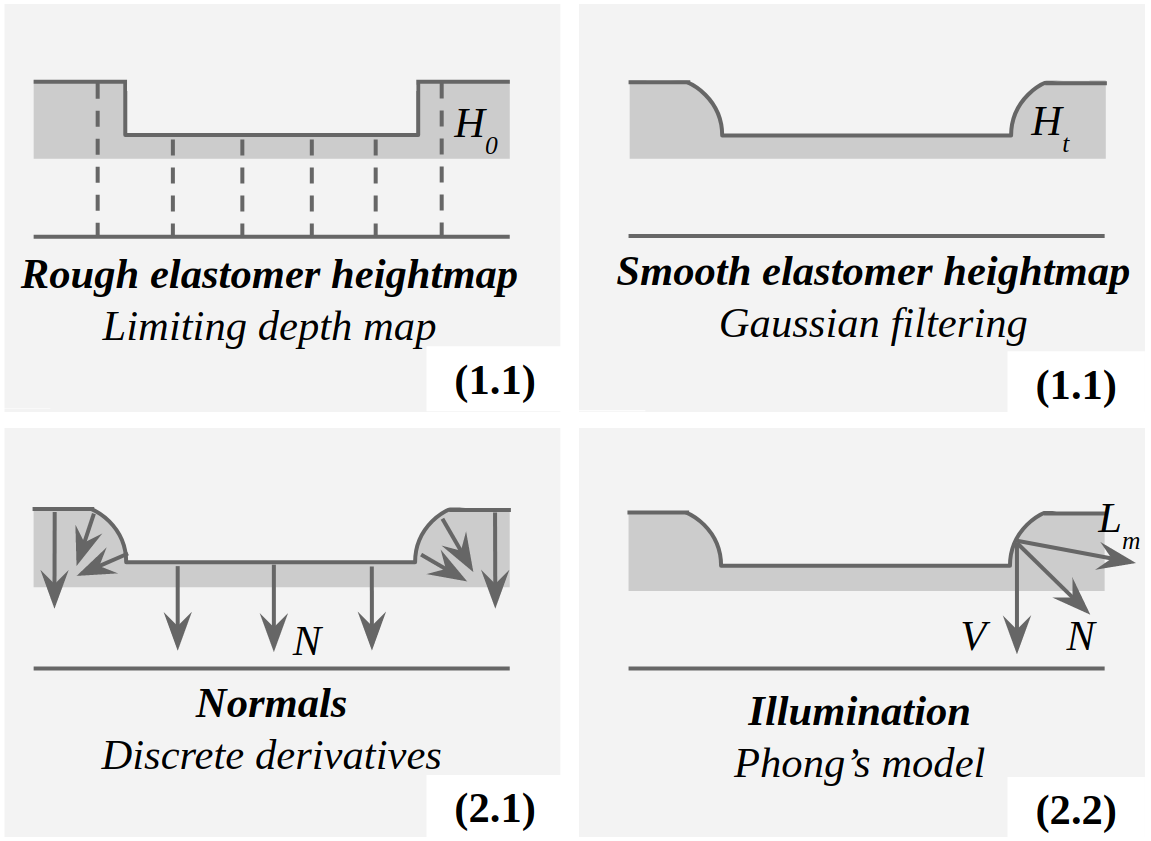}
\caption{The two steps in our proposed approach: 1) the elastomer heightmap is first approximated from a depth map captured by a depth camera, by \textbf{(1.1)} limiting the depth map and \textbf{(1.2)} smoothing it using Gaussian filtering; 2) then the elastomer internal illumination is rendered by \textbf{(2.1)} computing its surface normals as discrete derivatives and \textbf{(2.2)} applying Phong's illumination model.}
\label{fig:approach_steps}

\end{figure}

The GelSight sensors~\cite{RetrographicSensing} are built using a soft transparent membrane, coated with an opaque elastic paint and placed over a rigid transparent glass, as illustrated in Figure~\ref{fig:gelsight_working_principle}. When an object is pressed against the tactile membrane, the soft elastomer distorts and the geometry of the object is indented onto the elastomer. A view of the pressed surface can then be obtained by a camera enclosed within the opaque rigid shell. Light sources (LEDs) are placed inside the shell to illuminate the elastomer internal surface, making the sensor readings immune to external light variations. 

To facilitate the stereographic image processing, light from different colored LEDs is shone from different directions. In the GelSight (2014)~\cite{GelSightSmallParts}, four sets of LEDs (red, green, blue and white) are used, while in~\cite{GelSight2017GeometrySlip} only RGB LEDs are considered. Other variants such as \textit{GelSlim} \cite{GelSlim} are equipped only with white LEDs. As a result, different GelSight sensors produce different tactile images. In this work, we leverage the Phong's model to render the internal illumination, which can be \revised{parameterized to model any set of directional light sources}. Thanks to this, our approach can be trivially configured to simulate any GelSight-like sensor.

\section{The Proposed Simulation Model}

\label{sec:method}

As described in Section~\ref{sec:working_principle}, the core component of any GelSight-like sensor is its deformable tactile membrane that is internally illuminated by multi-color LEDs. We propose a novel approach to generate such tactile images directly from depth maps that can be easily captured in most of the off-the-shelf simulators. As illustrated in Figure~\ref{fig:approach_steps}, the proposed simulation model consists of two main steps: 1) the heightmap of the elastomer is first computed from the depth map of the object that is in contact with the elastomer; 2) the internal illumination of the elastomer is then computed using Phong's model~\cite{phongIlluminationModel}.


\subsection{The elastomer heightmap from the camera depth map}

\begin{figure}[t]
\centering
\includegraphics[width=0.49\textwidth]{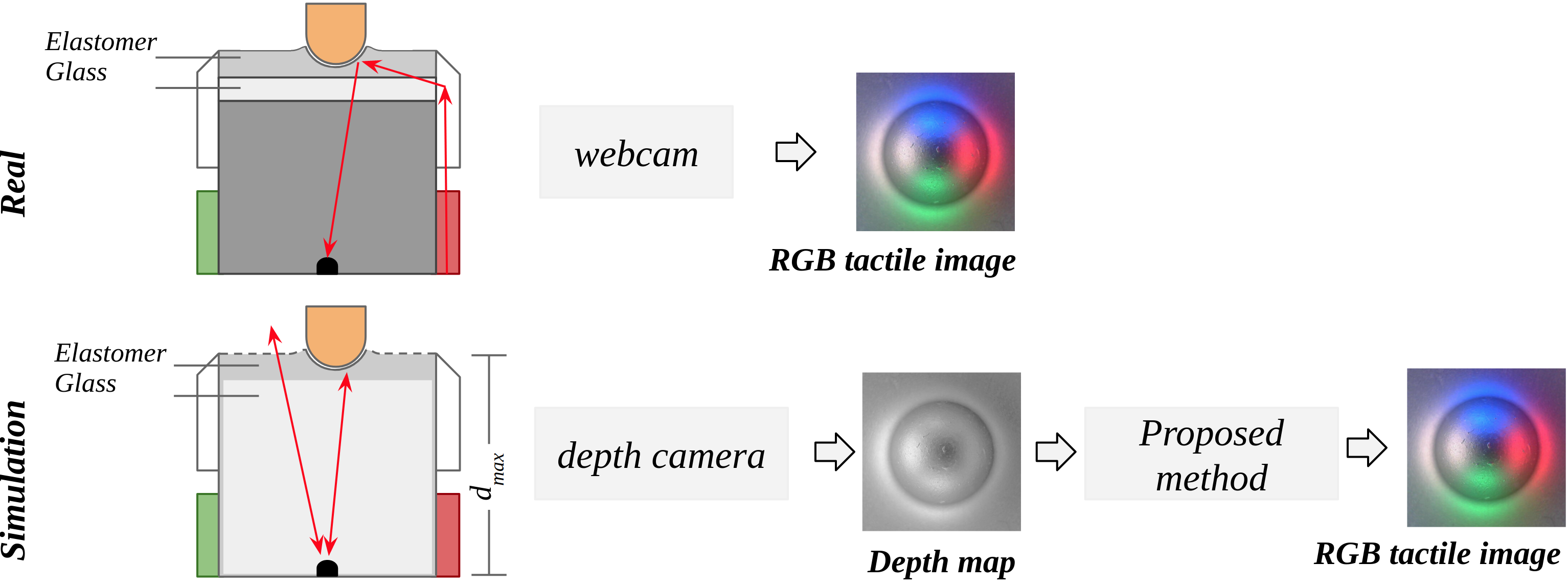}
\caption{\textbf{Top:} In the real GelSight tactile sensor, the webcam installed in its core directly captures the RGB tactile images. \textbf{Bottom:} In the proposed simulated GelSight sensor, a depth map is first captured by a depth camera, from which the virtual tactile image generated using the proposed simulation method. Note that in the simulated model, both the elastomer and glass are strategically placed such that they are invisible to the depth camera, as detailed in Section~\ref{subsec:virtual_setup}.}
\label{fig:sim_real_pipelines}
\end{figure}

To obtain the elastomer heightmap, a structured-light based depth camera is placed at the same position as the RGB camera in the real sensor, as shown in Figure~\ref{fig:sim_real_pipelines}. The simulated camera captures a depth image $D$ of the object in contact with the elastomer. The obtained depth map is then thresholded by the maximum distance $d_{max}$ to which the elastomer would be able to contact, resulting in the elastomer height map $H_0$:


\begin{equation} 
H_0(x,y) =   
\begin{cases} 
      D(x,y) & \text{if } D(x,y) \leq D_{max} \\
      d_{max} & \text{otherwise}
\end{cases}
\end{equation}
where $(x,y)$ is the location of a pixel in $D$.

$H_0$ captures the indentation caused by the object in contact with the elastomer, however, it contains sharp edges resulted from the thresholding. In contrast, the real elastomer \revised{presented} smoother edges inherent from its elastic properties that generate the color gradients around the in-contact indentation, as shown in Figure~\ref{fig:gelsight_working_principle}-D.
To approximate such smooth contact edges, without resorting to more computationally expensive algorithms, 2D (Bivariate) Gaussian filters $G(x,y)$ are applied incrementally over $H_0$:


\begin{align} 
G(x,y) &= \frac{1}{\sqrt{2\pi\sigma^2}} e^{-\frac{x^2+y^2}{2\sigma^2}} \\
H'_{t} &= H_{t-1} \ast G \label{eq:filter}\\
H_{t}(x,y) &=
\begin{cases} 
      H'_{t}(x,y)  & \text{if } H'_{t}(x,y) \geq H_{0}(x,y) \\
      H_{0}(x,y) & \text{otherwise}
\end{cases} \label{eq:merge}
\end{align}
where $\sigma$ is the standard deviation of $G$, $H_{t}$ is the elastomer surface approximation after $t \in [1, T]$  steps, $ \ast $ stands for the convolution operation. In each step: 1) $G$ is first applied to the elastomer surface approximation from the previous step $H_{t-1}$ to obtain the smoothed surface $H'_{t}$ (Equation~\ref{eq:filter}); 
2) $H'_{t}$ is merged with $H_0$ to preserve the sharp features within the in-contact region (Equation~\ref{eq:merge}).

To mimic the  bump  contouring  that  is  raised  around  the in-contact region by the depression caused by the contact force, a final heightmap is computed based on the difference of Gaussians.
\begin{align}
\label{eq:gaussians_subtraction}
    H_{DoG} &= 2 H_{narrow} - H_{wide} 
\end{align}
where $H_{narrow}$ is an heightmap approximation computed with a smaller $\sigma$ than the one used in $H_{wide}$. See Section~\ref{sec:elastomer_approximation} on the discussion about the elastomer deformation approximation variants.



\subsection{The internal illumination of the elastomer}

The generation of RGB tactile images $I$ from the heightmap $H$ of the elastomer can be interpreted as the inverse problem of the surface reconstruction problem~\cite{RetrographicSensing}, as the former consists on finding the mapping function $I \rightarrow H$ while the latter $H \rightarrow I$. In both cases, the relationship between the two can be described by:
\begin{equation}
\label{eq:eq_1}
I(x,y) =  R\left(\frac{\partial H}{\partial x}, \frac{\partial H}{\partial y}\right)
\end{equation}
where $R$ is the reflectance function that models both the lighting conditions (i.e., illumination of the LEDs) and the reflectance properties of the surface material (i.e., the elastomer coating paint). Here, it should be noted that the color observed at a given pixel is directly correlated with the orientation of the corresponding point on the elastomer.  

In~\cite{RetrographicSensing}, the mapping of the two points in the image space and the elastomer is built through a calibration process. In our case, we get $R$ using the Phong's illumination model~\cite{phongIlluminationModel}. Phong's model is an empirical model of local illumination that has been developed in the context of 3D Computer Graphics to describe how a given surface reflects light as a combination of the diffuse and specular reflections. 

From Phong's model, $I(x,y)$ observed at a given point of the sensor elastomer is given by three components: ambient, diffuse and specular light, as
\begin{align}
I(x,y) &=k_a i_a+\sum_{m \in L}{(k_d(\hat L_m \cdot \hat N) i_{m,d} + 
k_s(\hat R_m \cdot \hat V)^\alpha i_{m,s})} \label{eq:phong_model}\\
\hat R_m &= 2 (\hat L_m \cdot  \hat N ) \hat N - \hat L_m
\end{align}
where $L$ is the set of light sources (i.e., LEDs), $\hat L_m$ is the emission direction of a given light source $m$; $i_a$ is the intensity of the ambient light that is not caused by any of the LEDs; $i_{m,d}$ and $i_{m,s}$ are the intensities of the diffuse and specular reflections of light source $m$ respectively; $k_a$, $k_d$, $k_s$ and $\alpha$ are all reflectance properties of the surface; $\hat R_m$ is the direction that a perfectly reflected ray of the light would take; $\hat V$ is the direction pointing towards the camera. Given that our camera is pointing perpendicularly to the elastomer, $\hat V$ is set to $<0,0,1>$.
The surface normals $\hat N$ are computed using the discrete partial derivatives of the heightmap, as in the \revised{surface reconstruction}~\cite{RetrographicSensing}:
\begin{align} 
\hat N &= \;\; < \frac{\partial H}{\partial x}, \frac{\partial H}{\partial x}, -1 > \\
&= \;\; < \frac{H}{2r} \ast \begin{bmatrix}-1 &\!\!\!\! 0 &\!\!\!\! 1\end{bmatrix} , \frac{H}{2r} \ast  \begin{bmatrix}-1 &\!\!\!\! 0 &\!\!\!\! 1\end{bmatrix}^T, -1 > \nonumber
\end{align} 
where $r$ is a pixel-to-meter ratio obtained through a basic calibration process, detailed in Section~\ref{sec:virtual_datasets_and_parameters_setup}.


\section{Experimental Setup}
\label{sec:experimental_setup}

\begin{figure}
\centering
\includegraphics[width=0.49\textwidth]{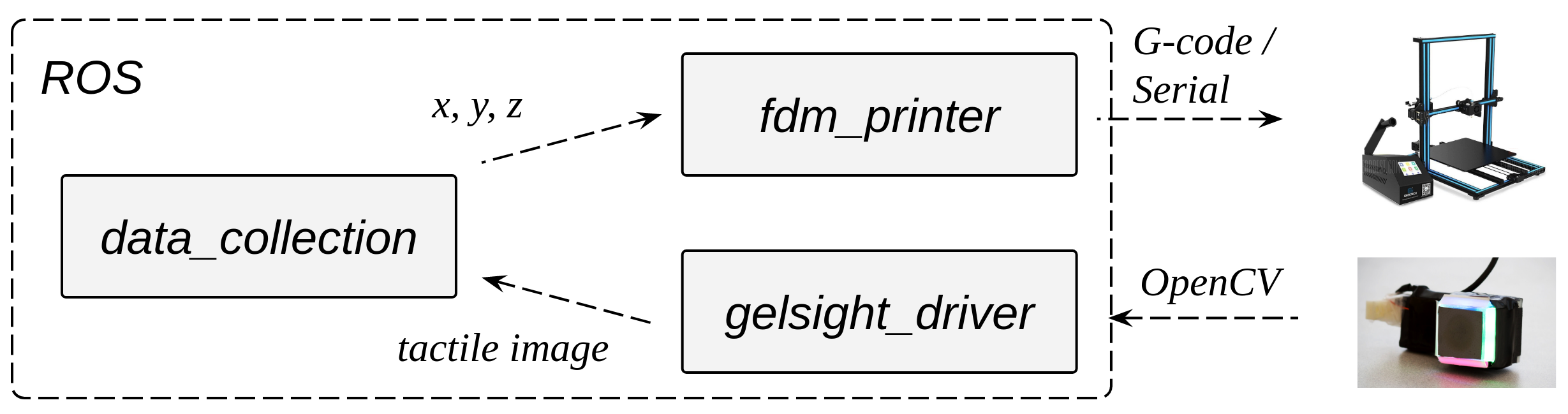}
\caption{To collect the (real) reference dataset, the \textbf{\textit{data\_collection}} node publishes position commands to the \textbf{\textit{fdm\_printer}} node while receiving tactile images from the \textbf{\textit{gelsight\_driver}} node. In turn, the \textbf{\textit{fdm\_printer}} node issues G\=/code commands to the 3D printer via serial communication, while the \textbf{\textit{gelsight\_driver}} node uses the \textit{OpenCV} library to capture the images from the sensor's webcam.}
\label{fig:data_collection_setup}
\end{figure}

\begin{figure}
\centering
\includegraphics[width=0.49\textwidth]{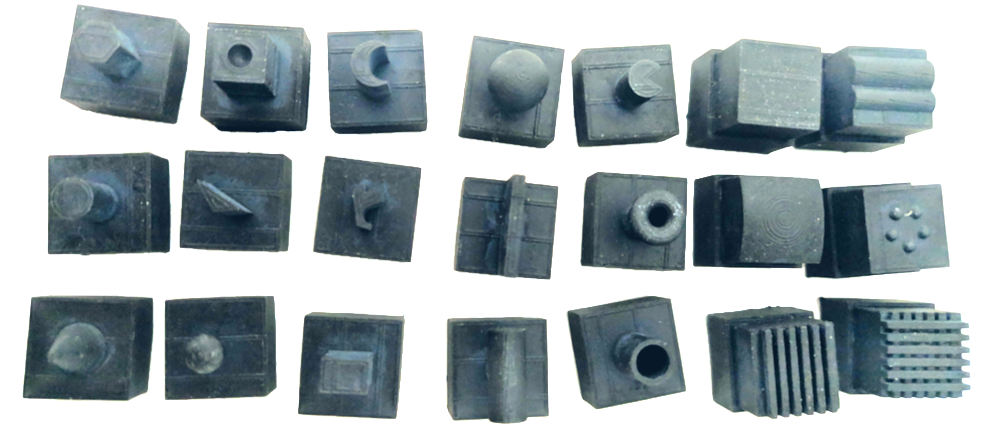}
\caption{The objects set. 1st row: Hexagon, Dot-in, Moon, Large Sphere, Pacman, Flat Slab, Wave; 2nd row: Cylinder, Triangle, Random Prism, Line, Torus, Curved Surface, Dots; 3rd row: Cone, Small Sphere, Rectangular Prism, Side Cylinder, Open Shell, Parallel lines and Crossed Lines.}
\label{fig:objects_set}
\end{figure}

\subsection{Real World setup}
\label{subsec:real_setup}
To produce the necessary real reference dataset, a GelSight sensor~\cite{GelSightSmallParts} is mounted onto a Fused Deposition Modeling (FDM) 3D printer \textit{A30} from Geeetech. One object from a 3D printed object set is placed onto the 3D printer bed and the printer is set to tap the object. As shown in Figure~\ref{fig:cover}-A, the sensor is fixed to the Tool Center Point of the printer, using a customised 3D printed fixture. Another fixture is attached to the printer base to ensure that all objects stay in the same position during the dataset collection. 

To automate data collection using the 3D printer, we create three nodes in the Robot Operating System (ROS)~\cite{ROS}, i.e.,  \textit{data\_collection}, \textit{fdm\_printer} and \textit{gelsight\_driver}. The \textit{data\_collection} node orchestrates the data collection process by publishing to the \textit{fdm\_printer} node and subscribing to the \textit{gelsight\_driver} node, as illustrated in~\mbox{Figure~\ref{fig:data_collection_setup}}. The \textit{fdm\_printer} driver node is implemented within the \textit{ros\_control} framework and has a custom \textit{hardware\_interface} that sends \textit{G\=/code} commands to control the 3D printer via serial communication. The \textit{gelsight\_driver} is implemented as a vanilla ROS publisher node, using the OpenCV library to interface with the webcam in the tactile sensor.


\subsection{Virtual World setup}
\label{subsec:virtual_setup}
Similar to the \textit{Real World} setup, the \textit{Virtual World} setup is also comprised of a simulated FDM printer, a simulated GelSight sensor and a set of virtual objects, as shown in \mbox{Figure~\ref{fig:cover}-B}. The entire setup is run in the \textit{Gazebo} simulator with its default Bullet physics engine. We choose \textit{Gazebo} as it is widely in robotics applications and can be easily integrated with ROS. However, since our approach only requires a simulated depth camera, it can be easily adapted to any other simulators that offer support to depth cameras or the lower level \textit{raycasting} operations.

The virtual 3D printer is modeled after the real one, using the Unified Robot Description Format (URDF). We adapt the mesh of the 3D printer to be its rigid links: the bed (x-axis), the y-axis link and the frame (z-axis). For the joints (or actuators), the \textit{ros\_control\_gazebo\_plugin} is used. From the printer's URDF description, the plugin set up the appropriate \textit{hardware\_interface} that establishes the communication between \textit{Gazebo} and \textit{ROS} via the \textit{ros\_control} framework.

The GelSight sensor is also modeled using the URDF specification. In this case, the mesh used to 3D print the real sensor shell is also used in the specification, together with six extra elements to model the sensor, i.e., the glass, the elastomer and four light guiding plates. The \textit{Gazebo}'s \textit{Openni Kinect} plugin is used to install the virtual depth camera at the core of the sensor. As detailed in Section~\ref{sec:method}, our proposed approach is used to generate the RGB tactile images from the obtained depth maps from the depth camera. 

We strategically set the glass and the elastomer as two concentric volumes, with the glass being placed inside the elastomer, as in the real sensor shown in Figure~\ref{fig:gelsight_working_principle} (Simulation). This ensures that the depth camera sensor sits inside these two volumes so that they are not \textit{hit} by the rays cast by the depth camera plugin, making them invisible in the depth maps. This is particularly necessary for the elastomer that needs to be opaque to look similar to the one on the real sensor. The elastomer is modeled without collision/physical geometry, to allow for objects to penetrate; in contrast, the glass is modeled as collision/physics only, to prevent any object from entering the rigid region of the sensor. The set of objects used in the simulation are the meshes of the 3D printed objects in the Real World setup.

\subsection{Reference (real) dataset}

\begin{figure}
\centering
\includegraphics[width=0.45\textwidth]{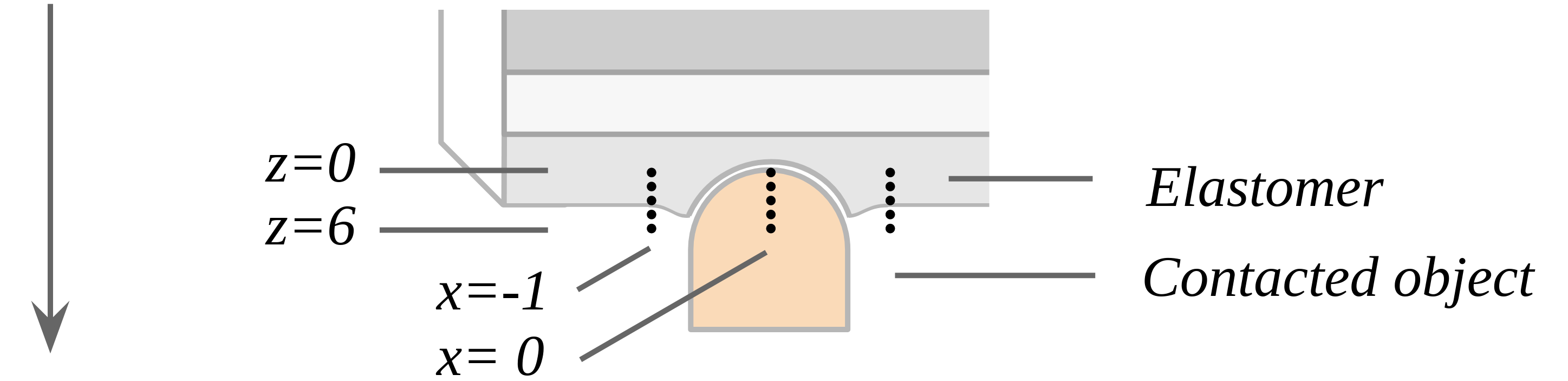}
\caption{\revised{2D illustration of the data collection motion. The GelSight sensor installed on the 3D printer is moved from top to bottom, contacting the object in a grid pattern and at multiple heights.}}
\label{fig:data_collection_motion}
\end{figure}

The reference (real) dataset consists of tactile images collected in normal contacts against $21$ 3D printed objects, shown in Figure~\ref{fig:objects_set}. The objects were printed using a Stereolithography (SLA) 3D printer \textit{Form 2} from the Formlabs and have different shapes on the top. Each object has a maximum volume of $1\times1\times2$ cm$^3$.

The contacts are located in a 3D grid of $3\times3\times11$, captured in \SI{1}{mm} horizontal steps and \SI{0.1}{mm} vertical steps, as shown in Figure~\ref{fig:data_collection_motion}. The grid is horizontally centered with the object, and its highest position ($z=0$) is where the very first contact is established in a top-down motion. This results in $99$ triplets of \mbox{\textit{ \textless~RGB Image, Class, Position \textgreater}} for each object. \revised{Each raw tactile sample has a resolution of $640\times480$.}

\subsection{Virtual datasets and baseline parameters setup}
\label{sec:virtual_datasets_and_parameters_setup}
A virtual dataset is also collected using a similar procedure to the reference dataset collection. However, in this case, depth maps $D$ are captured instead, to enable offline fine-tuning of the simulation method parameters. The corresponding synthetic tactile images are then generated by setting the method parameters manually. As in the real sensor, \revised{the captured depth-maps have a spatial resolution of $640\times480$.}

To mimic the real GelSight sensor, there are four light sources ($L$): white (top), blue (right), red (bottom) and green (left), as shown in \mbox{Figure~\ref{fig:cover}-C}. The color emitted by each light source is sampled from corresponding bright regions in a real tactile image, using the graphics editor GIMP. $k_d$ and $k_s$ of each LED are set based on the observed brightness of the corresponding LEDs. The ambient component $i_a$ is sampled from the corresponding position of a background image captured by the real tactile sensor, i.e., a tactile image captured when the sensor is not in contact with any surface; $k_a$ is set to $0.8$. All the obtained light configurations are listed in Table~\ref{table:lights_configs}. To obtain $r$, a cube of side \SI{5}{\mm} is placed near the virtual tactile sensor, and the distance (in pixels) between the first and last in-contact points of the same row is measured. 
The approximation parameters of the elastomer deformation are set based on visual inspection: the Bivariate Gaussian Kernel has a size of $21\times21$ and $\sigma$ is set to $7$; a total of $T=6$ steps are carried out. The maximum observable depth $d_{max}$ of the sensor is set to \SI{3}{\cm}, based on the real sensor dimensions, which corresponds to sum of the distance between the webcam and the elastomer (\SI{2.6}{\cm}) and the elastomer thickness (\SI{0.4}{\cm}).

\begin{table}
\centering

\caption{Baseline configurations for the virtual LEDs}
\def\arraystretch{1.2}
\begin{tabular}{  l | l  c  c  c } 
 & $i_{m,s}$, $i_{m,d}$ \tiny{(RGB)} & $\hat L_m$ \tiny{(XYZ)} & $k_d$ & $k_s$  \\
\hline
White & (255, 255, 255) & (0, 1, 0.25) & 0.6 & 0.5 \\
Blue & (115, 130, 255) & (-1, 0, 0.25) & 0.5 & 0.3 \\
Red & (225, 82, 108) & (0, -1, 0.25) & 0.6 & 0.4 \\ 
Green & (153, 255, 120) & (1, 0, 0.25) & 0.1 & 0.1 \\
\end{tabular}
\label{table:lights_configs}
\end{table}

All the materials used in this paper are available at https://danfergo.github.io/gelsight-simulation. This includes the packages for controlling the real and virtual setups, the collected datasets (both in the real world and in simulation), and the \textit{STL} files for the set of objects and fixtures.



\section{Experimental Results and Evaluation}
\label{sec:experiments}

\begin{figure*}[t]
\centering
\includegraphics[width=0.9\textwidth]{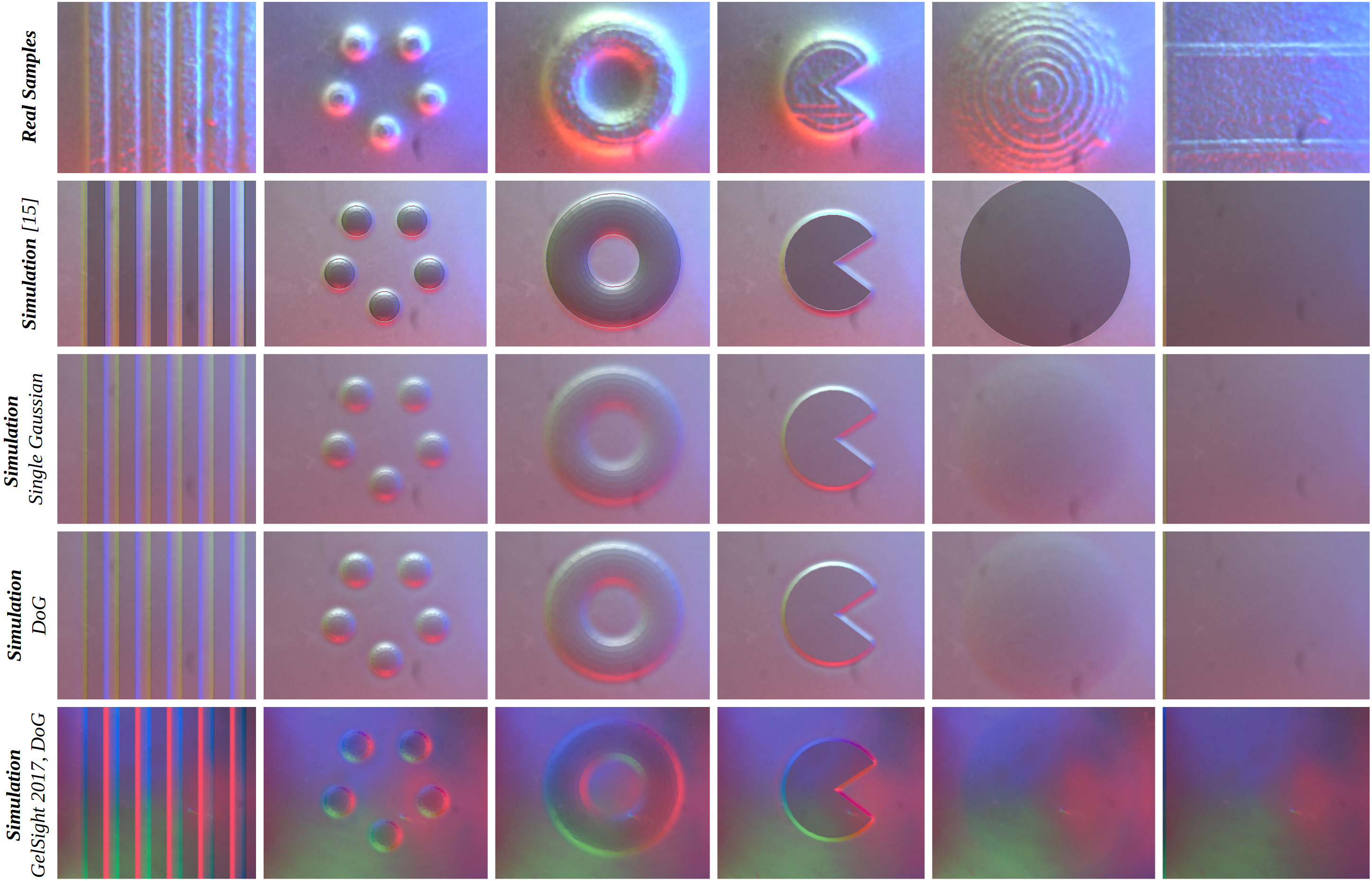}
\caption{Samples collected using a GelSight 2014 sensor (top row) and the corresponding simulations: using \cite{gomesgelsight} (2nd row), the ``Single Gaussian'' (3rd row) and ``Difference of Gaussians'' (4th row) for the elastomer heightmap approximation, for a GelSight 2017 \cite{dong2017improved} sensor (last row). As seen in the listed tactile images, the generated samples look realistic and quite similar to the real ones, being able to replicate internal light configurations of different sensors. For instance, the generated tactile images of the GelSight 2017 blends very well with the background image captured using the real sensor. On the other hand, some differences can be seen in the emitted reflections on areas with squashed, such as on the Torus and the Round Surface shapes.
}
\label{fig:samples}
\end{figure*}


\begin{table}
\centering
\caption{Real and generated datasets comparison}
\def\arraystretch{1.2}
\begin{tabular}{  l | r r r } 
 & SSIM & PSNR & MAE \\
\hline
Unaligned   & $0.731\pm0.005$ & $\bm{18.85\pm3.43}$ & ${\bm{8.40\pm0.04}}$\%  \\
Global Align & $0.852\pm0.009$ & $18.37\pm3.58$ & ${8.80\pm0.05}$\%  \\
Object Align & $\bm{0.859\pm0.004}$ & $18.58\pm3.04$ & $8.56\pm0.04$\%  \\ 
\hline
\cite{gomesgelsight} & $0.826\pm0.009$ & $17.58\pm6.32$ & $10.78\pm0.01$\%  \\

\end{tabular}
\label{table:alignment_errors}
\end{table}

We evaluate the proposed approach with three sets of experiments. Firstly, we compare the generated tactile images against corresponding samples collected from the real sensor, with both qualitative and quantitative analyses; and then, we demonstrate the use of our proposed simulated GelSight sensor in Sim2Real learning for a tactile classification task.

As shown in Figure~\ref{fig:samples}, the generated tactile images look very realistic and quite similar to the real samples collected using a real GelSight sensor, being able to replicate internal light configurations of different sensors.

\begin{figure}
\centering
\includegraphics[width=0.48\textwidth]{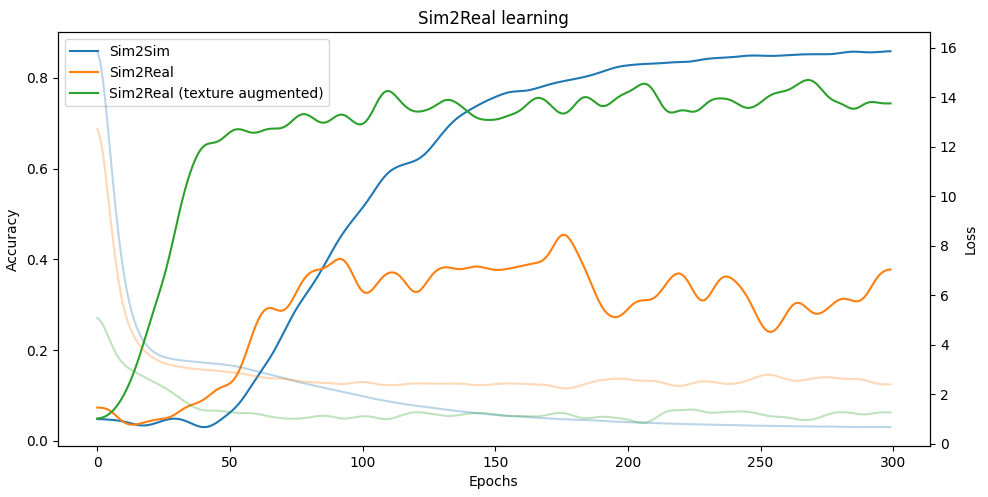}
\caption{The evolution of the accuracy and loss values, on the real validation split, while the network is being optimized using synthetic data. As shown, with the texture augmentation scheme described in \ref{sec:sim2real}, the proposed simulation method can be used for \textit{Sim2Real} transfer learning.}
\label{fig:tranfer_learning_losses}
\end{figure}

\subsection{Alignment of real and generated samples}
As we see in the previous sections, the settings \textit{Virtual World} and the \textit{Real World}, i.e., the trajectories executed to collect the real and generated tactile samples, are identical. To this end, the relative movement between two consecutive contacts is expected to be the same for both setups. However, due to the inaccurate placement of the objects onto the 3D printer bed center (in the \textit{Real World} setup), a small positioning error exists between both setups. A scaling error also exists due to inaccuracies in the virtual camera configurations, e.g., the distance from the sensor tactile membrane to the camera and the lens viewing angle. 

To mitigate these errors and to ensure a more accurate comparison, we start by using a single alignment pair of the real and virtual tactile images (by positioning the \textit{Dot-in} shape at \mbox{$<0,0,8>$}) to compute the transformation of the translation and the scaling between the two datasets. Two pairs of corresponding points in the real and virtual frames are selected and constrained to fall in two corresponding vertical lines. A third pair of points is also derived such that the 3 points in each frame form a right isosceles triangle. This constrained formation ensures that the alignment transformation produces only the desired translation and scaling transformation. The OpenCV \textit{getAffineTransform} and \textit{warpAffine} functions are then applied to find the Affine Transformation between the two frames and then the entire dataset. Due to the alignment operation, the transformed (real) frame has a smaller area than its original. As such, both images are cropped to their common area.

Three evaluation metrics are used to evaluate the alignment of the tactile images in the real dataset and the generated dataset: Structural Similarity (SSIM)~\cite{ssim}, Peak Signal-to-Noise Ratio (PSNR) and Mean Absolute Error (MAE). As shown in Table~\ref{table:alignment_errors}, in the Unaligned method, an average SSIM of 0.731 and an average MAE of 8.395\% are obtained, while the Aligned (single transform) method obtains a significantly improved SSIM 0.859 but a slightly worse MAE of 8.795\%.

By observing the different frames between aligned and non-aligned pairs, we find that other objects are not aligned as well as the \textit{Dot-in} object. To improve the alignment for each object, a second alignment approach is performed where the Affine transformation is instead used, and it is applied per each each object, \revised{resulting in aligned tactile samples with a resolution of, on average, $577.15\times455.30$}.  With this Alignment (per object) approach, a minimal improvement is obtained for both the SSIM (0.059) and the MAE (8.564\%), compared to the Aligned (single transform) approach, but the MAE is still higher than that of the Unaligned method (8.395\%). A probable reason for this is that the affine transformation aligns well for the salient features in the images, achieving a better SSIM, but introduces more distortions to the images at the same time, increasing the absolute errors.

\subsection{Analysis of the elastomer heightmap approximation}
\label{sec:elastomer_approximation}

\begin{figure}
\centering
\includegraphics[width=0.49\textwidth]{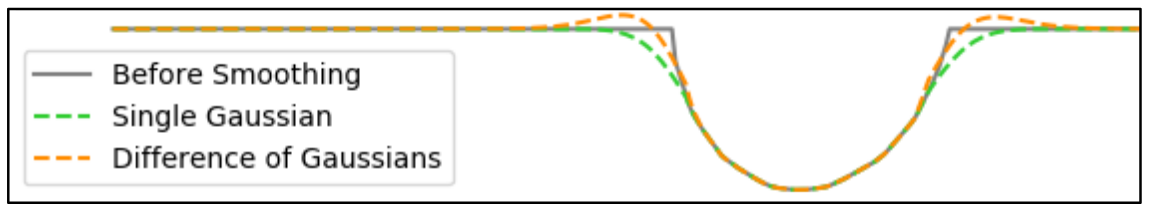}
\caption{Comparison of different methods for approximating elastomer deformations: without any smoothing effects (``Before Smoothing''), smoothed with a single Gaussian filter (``Single Gaussian'') and smoothed with the DoG (``Difference of Gaussians'').  The plotted cross section is extracted from part of an heightmap of the virtual elastomer pressing against the \textit{Dots} object.}
\label{fig:heightmap}
\end{figure}

When approximating the smooth curvature of the sensor elastomer using Gaussian filtering, one issue is that this process also smoothens the existing sharp features within the \textit{in-contact} regions. To mitigate this effect, the initial version of our approach, introduced in the workshop paper~\cite{gomesgelsight}, segments the tactile image into \textit{in-contact} ($H_0 < d_{max}$) and \textit{not-in-contact} ($H_0 = d_{max}$) regions. Then the smoothed and sharp (original) heightmaps are merged through element-wise multiplication, such that the Gaussian filtering only affects the \textit{not-in-contact} regions. However, in the produced tactile images, shown in \mbox{Figure~\ref{fig:samples}} (second row), a sharp contouring around the \textit{in-contact} regions can be noticed that do not reflect the behaviour of the real elastomer.

After extensive analysis, we conclude that this contour artifact is caused by the height discontinuity introduced by the two masks, and the incorrect assumption that elastomer contacts the in-contact object on its entire top-down 2D projection area. This observation leads to the replacement of these masks by the \textit{max} operations. A second issue in the single Gaussian approximation described in~\cite{gomesgelsight} is that it does not address the bump contouring around the \textit{in-contact} regions raised by the depression caused by the contact force. In this paper, we introduce the subtraction of two Gaussians, i.e., Difference of Gaussians, in Equation~\ref{eq:gaussians_subtraction}. The heightmaps generated by the ``Before Smoothing'', smoothed with ``Single Gaussian'' and smoothed with ``Difference of Gaussians'' methods are shown in Figure~\ref{fig:heightmap}. It can be seen in the figure that the Difference of Gaussians method approximates the elastomer deformation more accurately.

\subsection{Sim2Real transfer-learning for shape classification}
\label{sec:sim2real}

\begin{figure}
\centering
\includegraphics[width=0.49\textwidth]{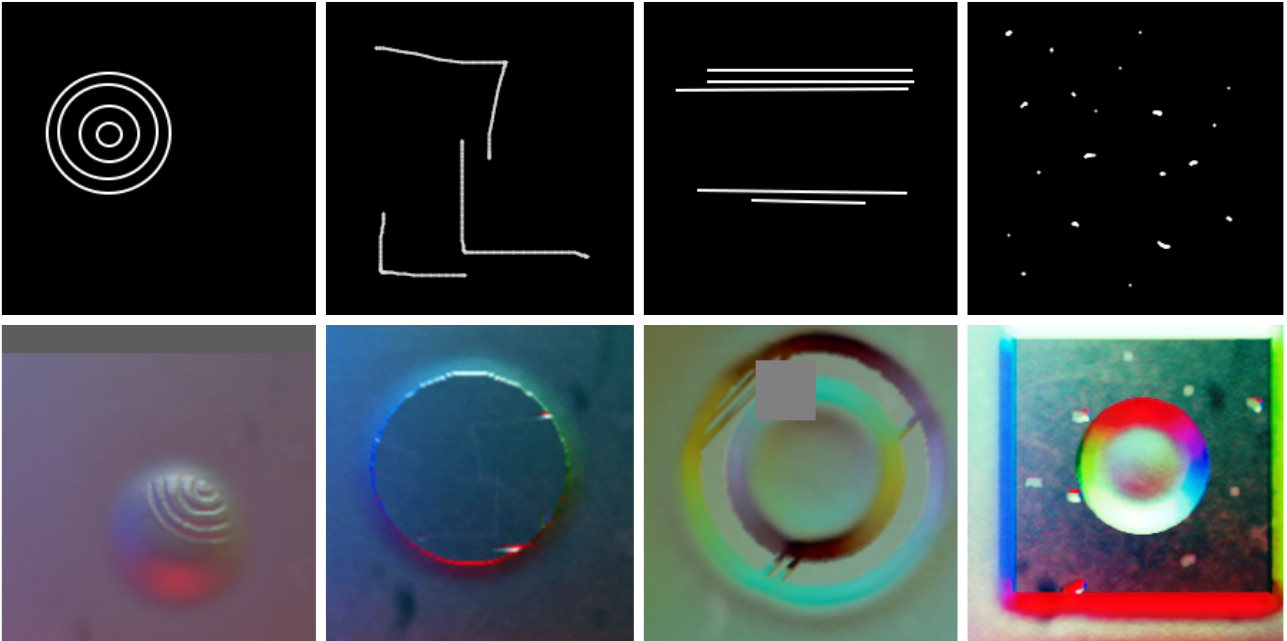}
\caption{On the top row, four of the twelve textures created to perturb the captured virtual depth-maps, to address the \textit{Sim2Real} gap.  On the bottom row, corresponding augmented samples fed to the Neural Network during training, after perturbing the depth-map  with the randomly distorted texture, generating the RGB tactile sample using the proposed method, and applying a random augmentation transformation. }
\label{fig:rnd_artifacts}
\end{figure}

In a second set of experiments, we study how models pre-trained using our proposed approach can be applied to real GelSight sensors, i.e., can be used for \textit{Sim2Real} transfer learning. For this goal, we choose a tactile object classification as a demonstrative task. In this case, given an image of an object in contact with the sensor, the goal is to predict what shape of the 21 objects it belongs to. The real and virtual dataset consist of 2,079 (21 $\times$ 99) samples each. Data splits are randomly defined for training, validation and test with, 70, 20 and 9 samples respectively, per each object.

\revised{One aspect to consider in the \textit{Sim2Real} learning is the \textit{Sim2Real} gap that results from characteristics of the real world being not modeled in the simulation. In our case, we find that one major difference between the real and synthetic samples are the textures introduced by the 3D printing process. To mitigate this issue, we create twelve texture maps using GIMP that resemble the textures observed in the real samples, as shown in Figure~\ref{fig:rnd_artifacts}. By randomly perturbing the captured virtual depth-maps with such textures, we are able to produce an effective data augmentation scheme that significantly improves the \textit{Sim2Real} transition, from a 43.76\% classification accuracy to 76.19\%, in the real data.}

To carry out the classification task, a Convolutional Neural Network (CNN) is implemented. The first convolutional layers are extracted from a ResNet-50 CNN~\cite{resnet}, pre-trained on the ImageNet~\cite{imagenet_cvpr09}, in order to ease the optimization process. \revised{These are then followed by two \mbox{\textit{128-d}} \textit{FC\=/ReLU\=/BatchNorm} blocks and a final \mbox{\textit{21-d}}  FC layer with a Softmax activation, which outputs the predicted object class as one-hot encoded vectors.} The network is optimized using the Adadelta optimizer, and a step size of $0.1$, by minimizing the categorical Cross-Entropy loss function. 

\revised{We report the classification accuracies in Table~\ref{table:sim2real_results}. For the \textit{Real2Real} (94.65\%) and \textit{Sim2Sim} (82.73\%) baselines, the CNN is trained and evaluated using the real and synthetic data-splits accordingly. For the \textit{Sim2Real} experiments, the CNN is trained using the training split of the synthetic dataset, and is evaluated on the validation and test splits of the real dataset. The validation results are used for assessing the training and choosing the best weights, while the test split is only used for final benchmarking. For all the experiments, the training samples are randomly augmented\footnote{https://github.com/aleju/imgaug}, and in the \textit{Sim2Real (texture augmented)} case, the synthetic tactile images are also dynamically generated from the depth-map that is perturbed by a randomly selected and randomly distorted texture, as shown in Figure~\ref{fig:rnd_artifacts}. As the network processes input images of $224\times224\times3$, during training and evaluation, the tactile images of variable sizes are square cropped and resized to fit the network.}




\begin{table}
\centering
\caption{Classification experiment results summary}
\def\arraystretch{1.2}
\begin{tabular}{  l | r r } 
 & Validation & Test \\
\hline
Real2Real                & 95.0\%  & 94.65\%  \\
Sim2Sim                  & 86.42\% & 82.73\%  \\
\hline
Sim2Real   & 43.80\% & 43.45\%  \\
Sim2Real (texture augmented) & \textbf{77.38}\% & \textbf{76.19}\%  \\

\end{tabular}
\label{table:sim2real_results}
\end{table}



\section{Conclusion and Discussion}
\label{sec:conclusions}

We introduce a novel way of generating tactile images from a simulated GelSight sensor, to enable \textit{Sim2Real} learning with high-resolution tactile sensing. Our proposed simulation method has been integrated with the widely used Gazebo simulator seamlessly. As the proposed method only depends on the surface function, it can be implemented in any other widely used robotics simulators. 

\revised{The proposed simulation approach is conceived to generate synthetic tactile images that would otherwise be captured by a flat GelSight sensor, as proposed in \cite{RetrographicSensing}. 
In \cite{RetrographicSensing}, lights are installed such that a direct mapping between the surface orientation and the captured pixel intensity exists.
As such, the assumptions in this work, i.e., a flat elastomer, directional light sources and no shadows, are sufficiently valid for sensors that are constructed following these working principles, such as the ones proposed in~\cite{GelSightSmallParts, dong2017improved}. More recently, other sensors have relaxed some of these constraints in favour of rounded surfaces \cite{gomes2020blocks} or increased measuring areas \cite{GelSlim}. As such, for each specific sensor different generalizations would have to be performed, even though the main pipeline, i.e., elastomer deformation modeling followed by illumination rendering, should still apply. Some GelSight sensors have also been equipped with elastomers containing printed markers. Such markers facilitate the tracking of the elastomer movements and the generation of corresponding applied force fields. In this work, we have not addressed the markers simulation, however, this has already been carried out in \cite{ding2020sim}, for the TacTip sensor. 
At this point, the GelSight sensors are an entire family with multiple variants, and this work represents the first attempt of simulating a GelSight sensor and application in the \textit{Sim2Real} learning context. Future works should seek to extend the proposed simulation method to the generation of tactile images for other \mbox{GelSight-like} sensors, such as the GelTip sensor \cite{gomes2020blocks}.
}

\bibliographystyle{IEEEtran}
\bibliography{references}

\end{document}